%% file: revision.tex
\documentclass[journal]{IEEEtran}

\ifCLASSINFOpdf
\else
   \usepackage[dvips]{graphicx}
\fi
\usepackage{url}

\usepackage{amsmath}
\usepackage{amsfonts}
\usepackage{graphicx}
\usepackage[colorlinks,
linkcolor = blue,
anchorcolor = blue,
citecolor = blue,
]{hyperref}
\usepackage{multirow}
\usepackage{pifont}
\usepackage{multicol}
\usepackage{makecell}
\usepackage{times}
\usepackage{epsfig}
\usepackage{multirow}
\usepackage{overpic}
\usepackage{marvosym}
\usepackage[capitalize]{cleveref}
\usepackage{xcolor}
\usepackage{xspace}
\usepackage{amsmath,amssymb}
\usepackage[accsupp]{axessibility}
\usepackage{threeparttable}

\usepackage{tikz}
\usetikzlibrary{spy}

\newcommand{\ie}{{\emph{i.e.}}\xspace}

\crefname{section}{Sec.}{Secs.}
\Crefname{section}{Section}{Sections}
\Crefname{table}{Table}{Tables}
\crefname{table}{Tab.}{Tabs.}

\begin{document}

\title{Learning Monocular Depth from Focus with \\ Event Focal Stack}
\author{Chenxu Jiang, Mingyuan Lin, Chi Zhang, Zhenghai Wang, and Lei Yu% <-this % stops a space
        
\thanks{This work was partially supported by the National Natural Science Foundation of China (62271354). Corresponding author: Lei Yu (ly.wd@whu.edu.cn).
% , and the Open Project Program of the Key Laboratory of Artificial Intelligence for Perception and Understanding, Liaoning Province (AIPU).
}
\thanks{C. Jiang, M. Lin, C. Zhang, and L. Yu are with the School of Electronic Information, Wuhan University, Wuhan, China.  
% X. Jia is affiliated with both the School of Artificial Intelligence and the School of Futher Technology, Dalian University of Technology, Dalian 116024, China. Y. Guo is with the Intelligent Science Technology Academy of CASIC, China.
% } %E-mail: \{wyg.eis, jiangcx, zhangchi1,ly.wd\}@whu.edu.cn. }
% \thanks{
Z. Wang is with the School of Information Engineering, Nanchang University, Nanchang, China. 
% Corresponding author: Lei Yu (ly.wd@whu.edu.cn).
}
% \thanks{Corresponding author: Lei Yu (ly.wd@whu.edu.cn).}
}

\markboth{IEEE SIGNAL PROCESSING LETTERS}{Jiang \MakeLowercase{\textit{et al.}}: Learning Monocular Depth from Focus with Event Focal Stack}
\maketitle

\begin{abstract}
Depth from Focus estimates depth by determining the moment of maximum focus from multiple shots at different focal distances, \ie the Focal Stack. 
% However, restricted by the limited sampling rate of conventional optical cameras, it is difficult to obtain sufficient focus cues during the focal sweep. 
However, the limited sampling rate of conventional optical cameras makes it difficult to obtain sufficient focus cues during the focal sweep. 
Inspired by biological vision, the event camera records intensity changes over time in extremely low latency, which provides more temporal information for focus time acquisition. In this study, we propose the EDFF Network to estimate sparse depth from the Event Focal Stack. Specifically, we utilize the event voxel grid to encode intensity change information and project event time surface into the depth domain to preserve per-pixel focal distance information. A Focal-Distance-guided Cross-Modal Attention Module is presented to fuse the information mentioned above. 
% A U-Net architecture is utilized to encode focus volume and predict coarse depth at each level. 
Additionally, we propose a Multi-level Depth Fusion Block designed to integrate results from each level of a UNet-like architecture and produce the final output.
% , effectively mitigating the impact of noise. 
Extensive experiments validate that our method outperforms existing state-of-the-art approaches.

\end{abstract}


\IEEEpeerreviewmaketitle

\section{Introduction}

\IEEEPARstart{D}{epth} from Focus / Defocus plays a crucial role in the realm of monocular 3D vision. It avoids the image matching process in depth estimation based on camera stereo disparity or motion parallax, reducing the additional resource consumption associated with acquiring synchronized multi-view images \cite{yang2022deep}.  Simultaneously, it addresses the limitation of traditional monocular depth estimation in estimating absolute depth \cite{bhoi2019monocular}. 
% This technique adheres to the principles of the thin-lens model and utilizes information obtained from changes in focus (Depth from Focus, DFF) or defocus blur (Depth from Defocus, DFD) to deduce the absolute depth of a scene. 
Previous research in Depth from Defocus (DFD) has focused on estimating the level of blurriness across images captured at varying focal distances, \ie the Focal Stack (FS), based on the assumption of a constant local depth and a specific distribution of the lens' Point Spread Function (PSF) \cite{pentland1987new}, \cite{subbarao1994depth}, \cite{zhuo2011defocus},  \cite{gur2019single}, \cite{si2023fully}. 
% This methodology allows for the inference of the scene's depth by analyzing the variations in blurriness attributed to different focal distances. 
However, the above assumptions may not always stay valid.
% in real-world scenarios. 
Conversely, Depth from Focus (DFF), which is not dependent on the prior of PSF, is determined by assessing the focus level at each point of images in FS, and considering the focal distance at the peak of focus as the absolute depth for that position \cite{grossmann1987depth}, \cite{ens1993investigation}, \cite{moeller2015variational}, \cite{suwajanakorn2015depth}, \cite{surh2017noise}, \cite{hazirbas2019deep}. However, the low sampling frequency of conventional optical cameras often makes it difficult for
% to ensure that an image with the highest relative degree of focus is captured at every location in the scene, and the error in the
determination of the focusing moments, leading to the inevitable degradation of depth estimation. 

Event cameras have recently been popular in various dynamic vision tasks \cite{chiavazza2023low}, \cite{furmonas2022analytical}, \cite{gasperini2023robust} \cite{liu2024event}, attributable to their asynchronous mechanism and high temporal resolution. Differing from traditional frame-based cameras, event cameras asynchronously record the intensity changes at each pixel when the log intensity change surpasses a predefined threshold \cite{gallego2020event}. These capabilities allow for more detailed and timely records during the focal sweep, producing the Event Focal Stack (EFS) \cite{lou2023all}, which provides more focus cues than traditional FS for depth estimation. \cite{haessig2019spiking} introduced the first event-based sparse DFD method, employing a couple of Leaky Integrate-and-Fire (LIF) neurons \cite{hunsberger2015spiking} to detect event polarity reversal time at each pixel, which indicates focus. However, the estimation results can be easily contaminated by noise owing to its event-by-event computing mechanism.

In this study, to address the aforementioned issues, we choose the more generalized DFF baseline and introduce the Event-based Depth from Focus (EDFF) Network that robust to event noise. 
% At the beginning, we revisit the fundamentals of DFF and why event is superior in this task in \cref{sec:Problem Formulation}.  
Initially, we revisit DFF fundamentals and explore the advantages of utilizing events in \cref{sec:Problem Formulation}.
Then we present the event representations utilized to encode intensity change and focal distance information while fully leveraging temporal information of events in \cref{sec:Event Representation for Rolling Shutter Deblurring}. 
And \cref{sec:Overall Architecture} shows our overall architecture.
We present a Focal-Distance-guided Cross-Modal (FDCM) Attention Module, designed to integrate features from events and corresponding focal distances. 
% Subsequently, we employ a U-Net to encode intensity-change information over time that encompasses the focus volume within the feature domain and thus robust to noise. 
Subsequently, we employ a UNet-like architecture to encode intensity-change information over time within the feature domain, capturing the focus volume and enhancing robustness to noise.
Ultimately, a Multi-level Depth Fusion Block (MDFB) receives the outputs from each level of the decoder and integrates them to produce the final depth, establishing a coarse-to-fine approach.
% to alleviate the impact of noise.
The main contributions of this study can be summarized in threefold:
\begin{itemize}
  \item [1)] 
 We present EDFF, a novel architecture designed for efficiently learning monocular sparse depth by utilizing high temporal resolution event focal stack.
  \item [2)]
  We employ attention mechanism on different modalities in FDCM and multi-scale coarse-to-fine approach in MDFB to achieve effective depth estimation. 
  \item [3)]
  We build two event focal stack datasets for event-based DFF with depth ground truth, on which we evaluate our method against the state-of-the-art frame-based methods.
\end{itemize}

\begin{figure}[htbp]
    \centering
    \includegraphics[width=1\linewidth]{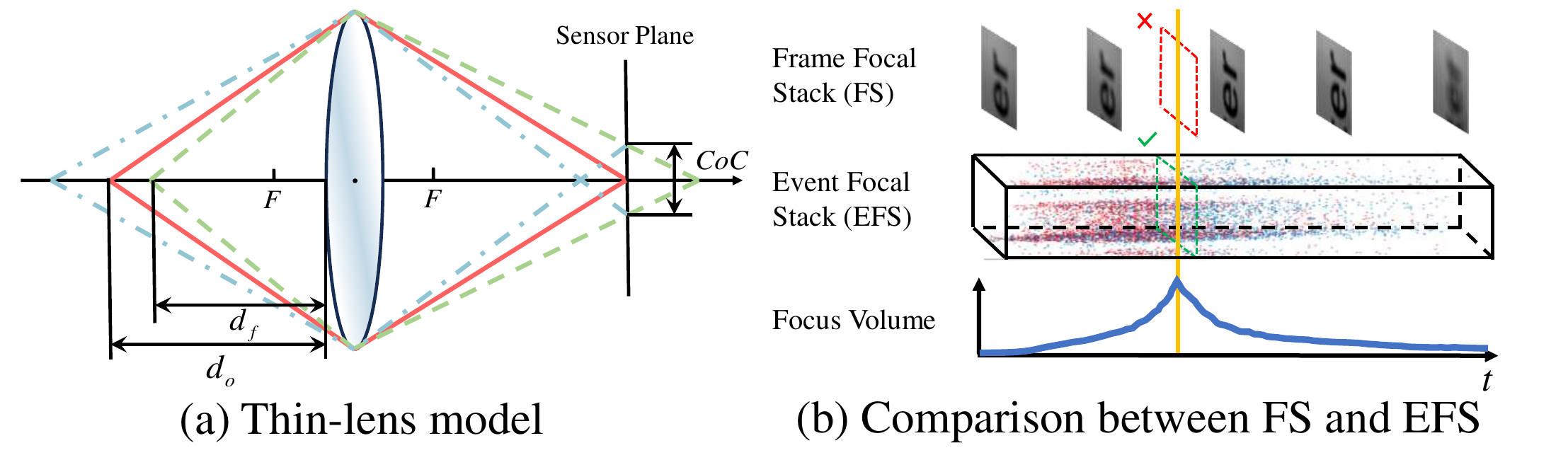}\vspace{-0.5em}
    \caption{(a) Illustration of the thin-lens model. (b) The maximum focus may be indiscernible in frame-based focal stacks due to the low sampling rate, which can be addressed by using an event-based focal stack with high temporal resolution.}
    
    \label{fig:thinlens}
    \vspace{-1.5em}
\end{figure}
\section{Methods}

\subsection{Problem Formulation}\label{sec:Problem Formulation}
We begin by reviewing the basic rules of DFF and the generation of EFS. Assuming that focus and defocus obey the thin-lens model illustrated in \cref{fig:thinlens}(a), one can derive \cite{gur2019single}:
\begin{equation}\label{eq:thin-lens model}
    s(\mathbf{x})=\frac{|d_f-d_o(\mathbf{x})|}{d_o(\mathbf{x})}\frac{F^2}{f\left(d_f-F\right)},
\end{equation}
where $F$, $f$ denote focal length and f-number of camera lens respectively, $d_f$ the focal distance, $d_o(\mathbf{x})$ the distance of target object at $\mathbf{x}\triangleq (x,y)$, in other words, depth. And $s(\mathbf{x})$ refers to the diameter of Circle of Confusion (CoC). The relationship between CoC and the amount of defocus can be expressed as \cite{pentland1987new}: 
\begin{equation}\label{eq:defocus generation}
    J(\mathbf{x})=I(\mathbf{x}) * G\big(r, \sigma^2(\mathbf{x})\big).
\end{equation}
In the above notation, $I(\mathbf{x})$ represents the All-in-Focus (AiF) image, and $G(r, \sigma^2(\mathbf{x}))$ the distribution of PSF, where $r$ is the location offset between $\mathbf{x}$ and the center of kernel $G$. $\sigma(\mathbf{x})$ is proportional to $s(\mathbf{x})$,
% \ie $\sigma(\mathbf{x})=\alpha s(\mathbf{x})$, 
and $J(\mathbf{x})$ refers to the image with defocus blur. $*$ denotes the convolution operator.
When a continuous sweep of focal distance is conducted, objects will preliminarily be defocus, then in focus gradually and defocus again. If the focus volume reach the maximum, $s(\mathbf{x})=0$ and $G$ degenerates into an impulse function, the current focal distance $d_f$ is then equal to the depth $d_o(\mathbf{x})$. That's how DFF works to evaluate the absolute depth. However, as shown in \cref{fig:thinlens}(b), the low sampling rate of traditional optical cameras may cause the focus point invisible. 

According to \cref{eq:defocus generation}, the change of CoC results in intensity variation, which can be sensed by event cameras at microsecond level and generate the EFS according to the event generation model:
\begin{equation}\label{eq:event generation model}
    \log\big({J}(\mathbf{x},t)\big)-\log\big({J}(\mathbf{x},\tau)\big) = p \cdot c,
\end{equation}
where $c$ is a predefined threshold, and $p\in \{-1,1\}$ denotes the polarity of events, indicating the direction of log intensity variation during the period $[\tau, t]$. The variation of intensity over time naturally encodes texture information, which contains critical focus details. Besides, due to the symmetry of the CoC in the vicinity of the focusing point, the event produces a polarity reversal before and after focusing \cite{haessig2019spiking}. This reversal point has been proved to be another significant pattern to distinguish by neural networks \cite{yang2022deep}.
So event camera is inherently suitable in DFF thanks to its high temporal resolution and implied intensity variation information. 
To fully leverage the characteristics mentioned above, alleviate the impact of event noise and achieve improved performance, we propose 
% to use event voxel grid and time surface as representations,
% , which will be interpreted in \cref{sec:Event Representation for Rolling Shutter Deblurring},
% and introduce 
a learning-based multi-scale architecture for event-based monocular sparse DFF.

\subsection{Event Representations}\label{sec:Event Representation for Rolling Shutter Deblurring}
% It's worth emphasizing that the texture variation, reversal of event polarity and time information are crucial for focus detection. 
% Instead of detecting event-by-event, we encode events into tensors and let the network learn focus information, alleviating the impact of event noise.
% Instead of detecting event-by-event, we encode a group of events into tensors to yield sufficient signal-to-noise ratio \cite{gallego2020event} and let the network learn focus information, alleviating the impact of event noise.
To mitigate the effects of event noise, we encode a group of events into tensors to yield a sufficient signal-to-noise ratio \cite{gallego2020event} and let the network learn focus cues.
Furthermore, the choice of event representation aims to characterize as much intensity variation, time, and polarity information as possible while incorporating focal distance information. Consequently, two distinct event representations are selected for this work.

\noindent{\bf Event Voxel Grid.} The event voxel grid is a commonly used input method in event-based deep learning 
% , as noted in 
\cite{liu2024event}, \cite{zhu2019unsupervised}, \cite{gehrig2019end},
which involves partitioning a set of $M$ events $\{e_k=(x_k,y_k,t_k,p_k)\}_{k\in [0,M-1]}$ into $N$ time bins and computing a weight decay for events, defined as:
% modify this representation by 
% A distinctive feature of it is the addition of a weight decay for events distant from the border of each bin, defined as:
\begin{equation}
    \mathcal{E}_V(\mathbf{x}, t_i) = \sum_{k} \delta(\mathbf{x}-\mathbf{x}_k)\max\big(0,1-|t_k-t_i|\big),
\end{equation}
where $t_i$ is scaled linearly from $t_0$ to $t_{M-1}$ across $N$ bins as $t_i=(N-1)(t_i-t_0)/(t_{M-1}-t_0)$, $i\in [1,N]$. 
We further separate the two polarity channels, resulting in a $N\times 2 \times H\times W$ tensor. The event voxel grid captures the intensity variations in detail and preserves temporal information, making it particularly suitable for this temporal accuracy-requiring task.

\noindent{\bf Event Time Surface.} In addition to intensity information, the focal distance during a focal sweep represents another critical piece of information needed to calculate absolute depth.
Assuming a linear relationship between focal distance and time, we first build time surface  \cite{zhu2018ev}, \cite{lagorce2016hots}, \cite{sironi2018hats} that encodes the timestamp of the most recent positive and negative events at each pixel, resulting in a tensor of dimensions $N\times 2 \times H\times W$. Then we project the time surface into the focal distance domain with a linear transformation, \ie
% it is common practice to uniformly sample $N$ focal distances for each bin. However, this method can result in a significant loss of time information due to the extensive time intervals associated with each bin. So 
% Thus we propose a new event representation called event time surface, which builds upon the concept of the event time surface \cite{zhu2018ev}, \cite{lagorce2016hots}, \cite{sironi2018hats}. 
% % the timestamp of an event can be projected onto the focal distance at which it is triggered. 
% % This concept underpins the utility of the  
% we divid the events into $N$ bins, 
% and encodes the timestamp of the most recent positive and negative events at each pixel,
% resulting in a tensor of dimensions $N\times 2 \times H\times W$. Assuming a linear relationship between focal distance and time, we then project the time surface into the focal distance domain, \ie
\begin{equation}
    \mathcal{E}_D(\mathbf{x}, t_i) = \operatorname{LT}_{t\rightarrow d_f}\big(\delta(\mathbf{x}-\mathbf{x}_j)\max(t_j)\big),
\end{equation}
where $t_j \in [t_{i},t_{i+1})$. In this way, it not only fully leverages the high temporal resolution of events but also provides per-pixel depth information with finer granularity.
We will refer to it as the 'Event Depth Surface' in the following text for convenience.
\begin{figure*}[t]
    \centering
    \includegraphics[width=0.98\linewidth]{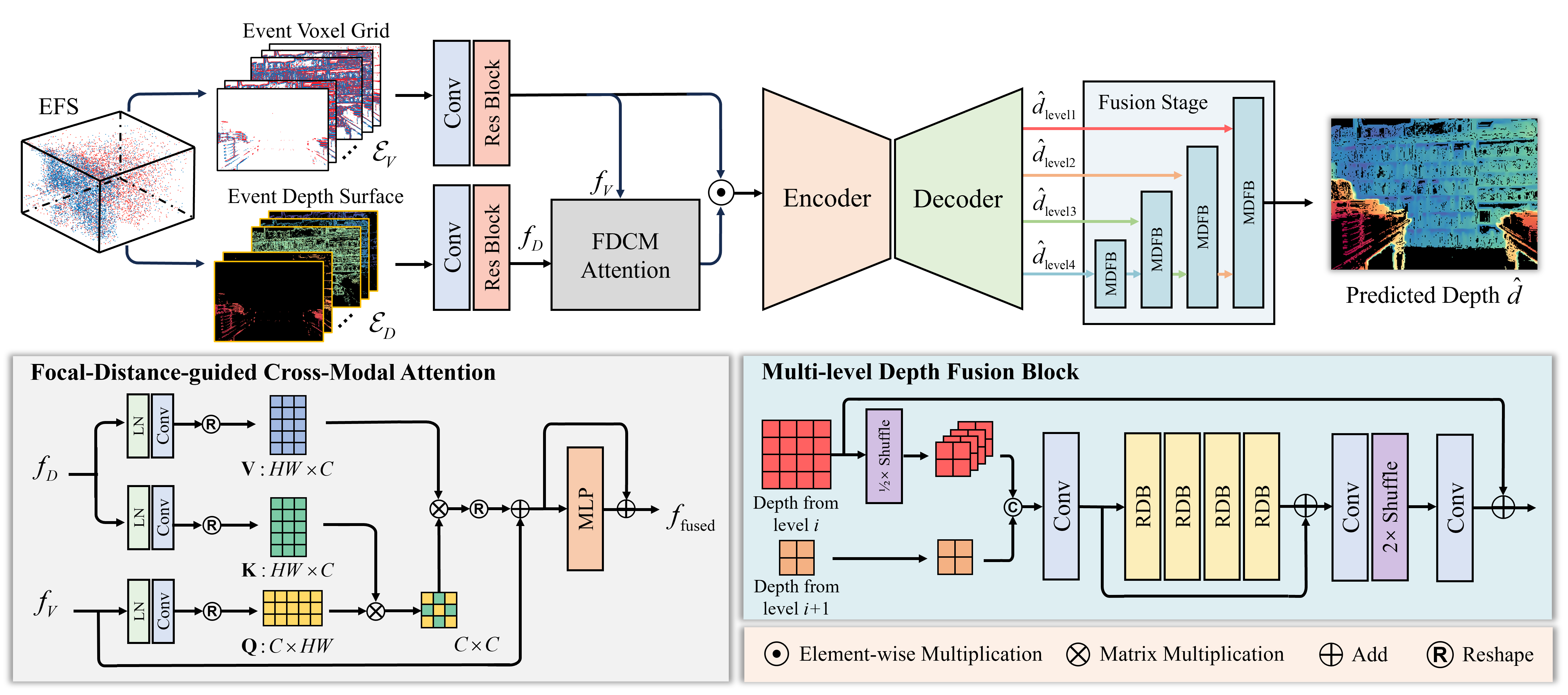}
    \vspace{-0.5em}
    \caption{ \textbf{Overview of EDFF.} A shallow feature extraction is adopted for each pair $\{\mathcal{E}_V,\mathcal{E}_D\}$. A FDCM attention is used to integrate features from the event domain $f_{V}$ and the depth domain $f_D$. The integrated features are then processed through a UNet-like architecture to thoroughly extract focus information and generate coarse results at multiple scales, which are visually differentiated using arrows of different colors. Finally, the output depths from various levels are fused using a MDFB to produce the final predicted depth.}
    \label{fig:overall}
    \vspace{-1.5em}
\end{figure*}

\subsection{Overall Architecture}\label{sec:Overall Architecture}

The overall architecture of the proposed EDFF is depicted in \cref{fig:overall}. The EFS traverses two distinct propagation paths with different representations. For each pair $\{\mathcal{E}_V,\mathcal{E}_D\}$, we begin with shallow feature extraction and get $\{f_V,f_D\}$. An FDCM attention module then integrates features from distinct intensity and depth modalities. The integrated features are subsequently processed through a UNet-like encoder and decoder to thoroughly analyze the changes in intensity and polarity, extract focus information, and generate coarse results at multiple scales. The output features from different levels are then fused using an MDFB to produce the final predicted depth, thus establishing a coarse-to-fine structure.

\noindent{\bf Focal-Distance-guided Cross-Modal Attention Module.} As each event corresponds to a specific focus distance, only those events near the focus time are deemed significant. 
% Therefore, jointly extracting features from events and depth information is critical for accurate absolute depth estimation. 
To address this, we introduce the Focal-Distance-guided Cross-Modal Attention Module, designed to compute a per-pixel weight to help learn the focal distance that most significantly impacts the final prediction. Specifically, we apply layer normalization and a convolutional layer to the voxel grid and depth surface to encode the query $\mathbf{Q}\in \mathbb{R}^{C\times HW}$ in the event domain, value $\mathbf{V}\in \mathbb{R}^{HW\times C}$ and key $\mathbf{K}\in \mathbb{R}^{HW\times C}$ in focal distance domain. The cross-modal attention is then computed as follows:
\begin{equation}\label{eq:attn}
    \operatorname{Attn}(\mathbf{Q},\mathbf{K},\mathbf{V})=\mathbf{V}\operatorname{Softmax}\bigg(\frac{\mathbf{Q}^\mathbf{T}\mathbf{K}}{\sqrt{d}}\bigg).
\end{equation}
A residual structure is incorporated during the attention computation to fully preserve texture and intensity change information that indicates focus from the event domain. Finally, the fused output $f_{\text{fused}}$ is computed using a Multilayer Perceptron (MLP) layer with a skip connection.

\noindent{\bf Multi-level Depth Fusion Block.} Enlightened by the multi-scale result recovery found in many frame-based DFF methods \cite{yang2022deep}, \cite{fujimura2023deep} which enhance robustness, we also employ a UNet-like architecture to generate results from different levels. However, we observe that in the above methods, only the output from the largest scale is used during the evaluation stage. The results from lower levels can serve as preliminary outcomes to guide the higher levels. Thus, we implement a layer-by-layer fusion strategy to produce a single, final result in a coarse-to-fine manner. 
% Since there is no extra supervision for the outputs of each decoder layer, these can be considered as features.
Drawing inspiration from the exceptional performance of the Residual Dense Block (RDB) \cite{zhang2018residual} in dealing with multi-scale scenarios, we propose a Multi-level Depth Fusion Block designed to learn the residuals between two adjacent layers. Specifically, the initial depth from level 
$i$ is reshuffled to a lower scale and concatenated with the initial depth from level $i+1$. A combination of multiple convolutional layers and RDBs is then used to fuse them. The fused result is reshuffled back to the upper scale, passed through a convolutional layer, and added to the initial depth of level $i$ to produce the output, which is then passed to the next fusion phase.

\subsection{Loss Function}\label{sec:Loss Function}
The loss function of the network is a combination of $L_2$ loss and smooth loss
with $\mathcal{M}(\cdot)$ used to mask out pixels that do not have events and $\alpha$, $\beta$ the balance parameters, \ie
\begin{equation}\label{loss}
    \mathcal{L} = \alpha\Vert\mathcal{M}(\hat{d})-\mathcal{M}(d)\Vert_2+\beta\Vert\mathcal{M}(\nabla\hat{d})-\mathcal{M}(\nabla d)\Vert_1.
\end{equation}

\section{Experiment}\label{Sec:Experiment}
\def\ssxxsone{(-0.25,-0.5)} % 小方框位置
\def\ssyysone{(-0.23,0.63)} % 大方框位置
\def\ssxxstwo{(-0.82,-0.5)} % 小方框位置
\def\ssyystwo{(0.23,-0.63)} % 大方框位置
\def\ssmag{2}
\def\ssmagtwo{1.8}
\def\ssizz{0.8cm} %框大小
\def\sswidth{0.115\textwidth} %子图大小
\begin{figure*}[t]
    \centering
    \begin{tabular}{c c c c c c c c}
        \begin{tikzpicture}[spy using outlines={green,magnification=\ssmag,size=\ssizz},inner sep=0]
    		\node {\includegraphics[width=\sswidth]{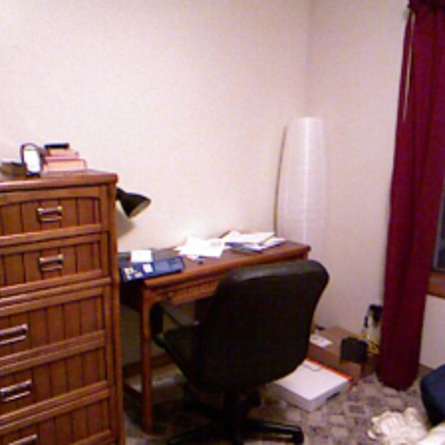}};
    		% \spy on \ssxxsone in node [left] at \ssyysone;
            % \spy [red] on \ssxxstwo in node [right,red] at \ssyystwo;
    	\end{tikzpicture} & \hspace{-4.5mm}
     \begin{tikzpicture}[spy using outlines={green,magnification=\ssmag,size=\ssizz},inner sep=0]
    		\node {\includegraphics[width=\sswidth]{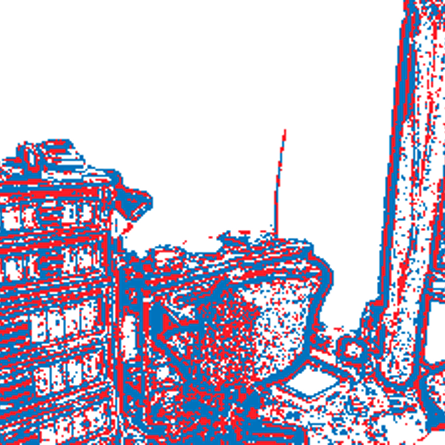}};
    		% \spy on \ssxxsone in node [left] at \ssyysone;
      %       \spy [red] on \ssxxstwo in node [right,red] at \ssyystwo;
    	\end{tikzpicture} & \hspace{-4.5mm}
        \begin{tikzpicture}[spy using outlines={green,magnification=\ssmag,size=\ssizz},inner sep=0]
    		\node {\includegraphics[width=\sswidth]{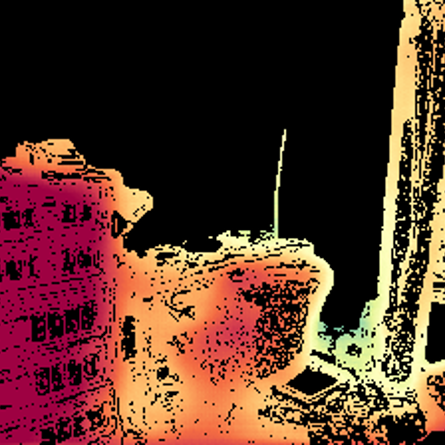}};
    		\spy on \ssxxsone in node [left] at \ssyysone;
      %       \spy [red] on \ssxxstwo in node [right,red] at \ssyystwo;
    	\end{tikzpicture} & \hspace{-4.5mm}
        \begin{tikzpicture}[spy using outlines={green,magnification=\ssmag,size=\ssizz},inner sep=0]
    		\node {\includegraphics[width=\sswidth]{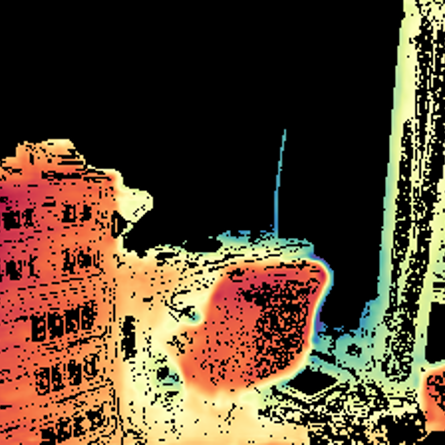}};
    		\spy on \ssxxsone in node [left] at \ssyysone;
      %       \spy [red] on \ssxxstwo in node [right,red] at \ssyystwo;
    	\end{tikzpicture} & \hspace{-4.5mm}
        \begin{tikzpicture}[spy using outlines={green,magnification=\ssmag,size=\ssizz},inner sep=0]
    		\node {\includegraphics[width=\sswidth]{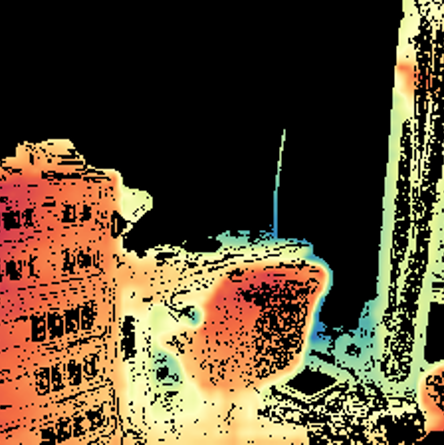}};
    		\spy on \ssxxsone in node [left] at \ssyysone;
      %       \spy [red] on \ssxxstwo in node [right,red] at \ssyystwo;
    	\end{tikzpicture} & \hspace{-4.5mm}
        \begin{tikzpicture}[spy using outlines={green,magnification=\ssmag,size=\ssizz},inner sep=0]
    		\node {\includegraphics[width=\sswidth]{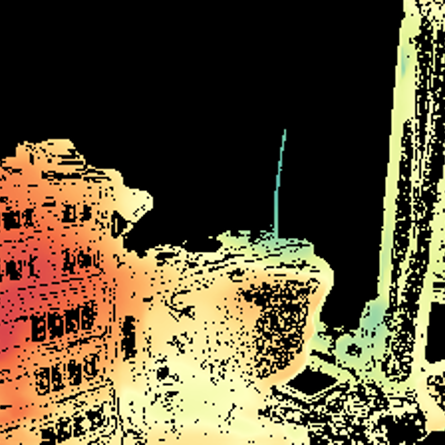}};
    		\spy on \ssxxsone in node [left] at \ssyysone;
      %       \spy [red] on \ssxxstwo in node [right,red] at \ssyystwo;
    	\end{tikzpicture} & \hspace{-4.5mm}
        \begin{tikzpicture}[spy using outlines={green,magnification=\ssmag,size=\ssizz},inner sep=0]
    		\node {\includegraphics[width=\sswidth]{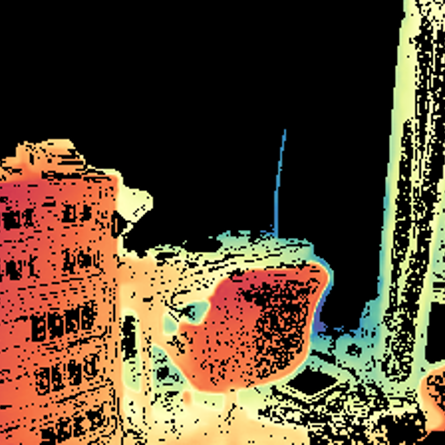}};
    		\spy on \ssxxsone in node [left] at \ssyysone;
      %       \spy [red] on \ssxxstwo in node [right,red] at \ssyystwo;
    	\end{tikzpicture} & \hspace{-4.5mm}
        \begin{tikzpicture}[spy using outlines={green,magnification=\ssmag,size=\ssizz},inner sep=0]
    		\node {\includegraphics[width=\sswidth]{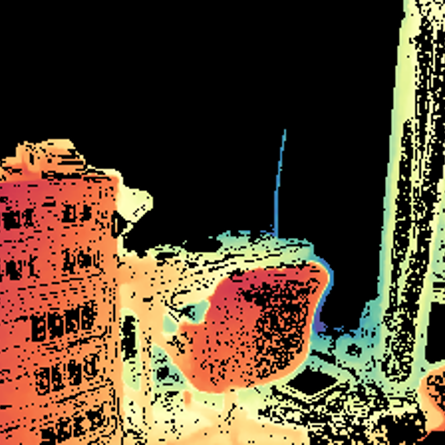}};
    		\spy on \ssxxsone in node [left] at \ssyysone;
      %       \spy [red] on \ssxxstwo in node [right,red] at \ssyystwo;
    	\end{tikzpicture}\vspace{-0.5em} \\
        \hspace{-3mm}\footnotesize{(a) AiF} & \hspace{-3mm}\footnotesize{(b) Events} & \hspace{-3mm}\footnotesize{(c) *DDFF} & \hspace{-3mm}\footnotesize{(d) *Defocus-Net} & \hspace{-3mm}\footnotesize{(e) *DFF-DFV}&
        \hspace{-3mm}\footnotesize{(f) DDFS}&
        \hspace{-3mm}\footnotesize{(g) EDFF (Ours)} &
        \hspace{-3mm}\footnotesize{(h) GT}\\
        \begin{tikzpicture}[spy using outlines={green,magnification=\ssmag,size=\ssizz},inner sep=0]
    		\node {\includegraphics[width=\sswidth]{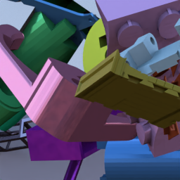}};
    		% \spy on \ssxxsone in node [left] at \ssyysone;
      %       \spy [red] on \ssxxstwo in node [right,red] at \ssyystwo;
    	\end{tikzpicture} & \hspace{-4.5mm}
        \begin{tikzpicture}[spy using outlines={green,magnification=\ssmag,size=\ssizz},inner sep=0]
    		\node {\includegraphics[width=\sswidth]{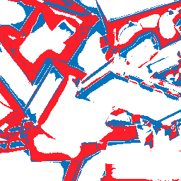}};
    		% \spy on \ssxxsone in node [left] at \ssyysone;
      %       \spy [red] on \ssxxstwo in node [right,red] at \ssyystwo;
    	\end{tikzpicture} & \hspace{-4.5mm}
        \begin{tikzpicture}[spy using outlines={green,magnification=\ssmagtwo,size=\ssizz},inner sep=0]
    		\node {\includegraphics[width=\sswidth]{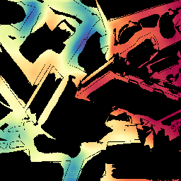}};
    		% \spy on \ssxxsone in node [left] at \ssyysone;
            \spy [red] on \ssxxstwo in node [right,red] at \ssyystwo;
    	\end{tikzpicture} & \hspace{-4.5mm}
        \begin{tikzpicture}[spy using outlines={green,magnification=\ssmagtwo,size=\ssizz},inner sep=0]
    		\node {\includegraphics[width=\sswidth]{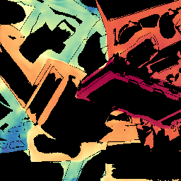}};
    		% \spy on \ssxxsone in node [left] at \ssyysone;
            \spy [red] on \ssxxstwo in node [right,red] at \ssyystwo;
    	\end{tikzpicture} & \hspace{-4.5mm}
        \begin{tikzpicture}[spy using outlines={green,magnification=\ssmagtwo,size=\ssizz},inner sep=0]
    		\node {\includegraphics[width=\sswidth]{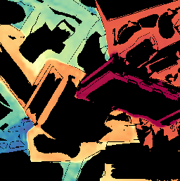}};
    		% \spy on \ssxxsone in node [left] at \ssyysone;
            \spy [red] on \ssxxstwo in node [right,red] at \ssyystwo;
    	\end{tikzpicture} & \hspace{-4.5mm}
        \begin{tikzpicture}[spy using outlines={green,magnification=\ssmagtwo,size=\ssizz},inner sep=0]
    		\node {\includegraphics[width=\sswidth]{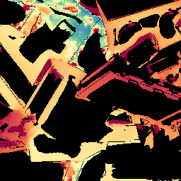}};
    		% \spy on \ssxxsone in node [left] at \ssyysone;
            \spy [red] on \ssxxstwo in node [right,red] at \ssyystwo;
    	\end{tikzpicture} & \hspace{-4.5mm}
        \begin{tikzpicture}[spy using outlines={green,magnification=\ssmagtwo,size=\ssizz},inner sep=0]
    		\node {\includegraphics[width=\sswidth]{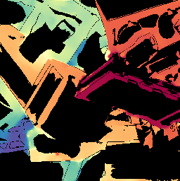}};
    		% \spy on \ssxxsone in node [left] at \ssyysone;
            \spy [red] on \ssxxstwo in node [right,red] at \ssyystwo;
    	\end{tikzpicture} & \hspace{-4.5mm}
     \begin{tikzpicture}[spy using outlines={green,magnification=\ssmagtwo,size=\ssizz},inner sep=0]
    		\node {\includegraphics[width=\sswidth]{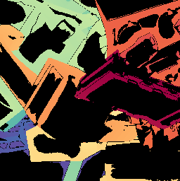}};
    		% \spy on \ssxxsone in node [left] at \ssyysone;
            \spy [red] on \ssxxstwo in node [right,red] at \ssyystwo;
    	\end{tikzpicture}\vspace{-0.5em}\\
        \hspace{-3mm}\footnotesize{(a) AiF} & \hspace{-3mm}\footnotesize{(b) Events} & \hspace{-3mm}\footnotesize{(c) *DDFF} & \hspace{-3mm}\footnotesize{(d) *Defocus-Net}& \hspace{-3mm}\footnotesize{(e) *DFF-DFV} &
        \hspace{-3mm}\footnotesize{(f) AiFDepthNet} & 
        \hspace{-3mm}\footnotesize{(g) EDFF (Ours)} & 
        \hspace{-3mm}\footnotesize{(h) GT} 
         \\
    \end{tabular}
    \caption{Qualitative comparison on the EFS-NYUv2 (top) and EFS-Blender (bottom) dataset. The warmer the color, the shallower the depth. We mark the retrained methods with * for identification. Details are zoomed in for a better view.}
    \vspace{-1.0em}
    \label{fig:result_vimeo}
    
\end{figure*}
\input{tables/quantitaitve}
\subsection{Datasets and Implementation}\label{sec:datasets and implementation}
Since no dataset exists for EFS-based depth estimation, we build two distinct synthetic datasets for training and evaluation. 
% In practice, we first render a frame-based focal stack with 480 images from two common DFF datasets.

\noindent{\bf EFS-NYUv2} is derived from NYUv2 \cite{silberman2012indoor}, an RGB-D dataset with real-world images and corresponding depth GT. The max depth is $9.99m$. We render the focal stack utilizing the PSF layer proposed in \cite{gur2019single}, with $F=50mm$, $f=8$. $d_f$ is uniformly distributed from $[1,10]m$ across 480 samples. 

\noindent{\bf EFS-Blender} is a synthetic dataset rendered by Blender using the code presented in \cite{zhuo2011defocus}. 
In our scene, 20 objects are randomly placed in front of a wall from $[1.5,10]m$, with a virtual optical camera shooting the defocused image.
The lens parameters are $F=25mm$, $f=1.2$, and $d_f\in [1,10]m$, with a total of 480 images sampled.

\noindent{Then}, we simulate the event focal stack with the rendered images using ESIM \cite{Rebecq18corl}.

Evaluation metrics include RMSE (measured in meters), AbsRel, $\delta<1.25$, $\delta<1.25^2$, and $\delta<1.25^3$.
We implement our network with PyTorch \cite{paszke2019pytorch} on an NVIDIA GeForce RTX 2080 GPU. The input EFS is randomly cropped into $200\times200$ patches. We utilize the Adam optimizer \cite{kingma2014adam} and the SGDR scheduler \cite{loshchilov2016sgdr}, with an initial learning rate of $5\times 10^{-4}$. The model is trained for 200 and 300 epochs, respectively, on EFS-NYUv2 and EFS-Blender datasets. The weighting factors $\alpha$ and $\beta$ in \cref{loss} are set to 128 and 1.

\subsection{Quantitative and Qualitative Evaluation}\label{Quantitative Evaluation}
As shown in \cref{tab:quantative}, we compare our EDFF against several frame-based supervised learning methods for DFF or DFD, including DDFF \cite{hazirbas2019deep}, Defocus-Net \cite{maximov2020focus}, AiFDepthNet \cite{wang2021bridging}, DFF-DFV \cite{yang2022deep}, and DDFS \cite{fujimura2023deep}. We apply the mask $\mathcal{M}$ to each method and compare sparse results. Retrained methods are labeled with *. 
The results for AiFDepthNet on EFS-NYUv2, DDFS on EFS-Blender, and the event-based method \cite{haessig2019spiking} are not presented because their training or testing code has not been released, and we rely on the pre-trained models to obtain the current results. 
% The results for AiFDepthNet and DDFS are not presented or retrained because their training code has not been released. 
% Note that existing DFF methods often select data with a small depth range or rescale it to primarily fall within $[0,3]m$ \cite{fujimura2023deep} because DFFs works better on short ranges \cite{maximov2020focus}. In contrast, our datasets cover ranges that are several times as large as theirs. 
Our EDFF significantly outperforms state-of-the-art methods on the EFS-NYUv2 dataset. According to our statistics, the maximum depth of NYUv2 varies from $[2,10]m$. Thus, uniform sampling of FS from $[0,10]m$ may yield less valid focus information for data in the shorter depth range. 
This issue is effectively addressed by utilizing the events, which feature high time resolution and capture detailed intensity change information, providing enhanced focus cues. Consequently, an EFS designed to focus from $[0,10]m$ can adeptly handle any depth range within this scope. In contrast, the EFS-Blender dataset tends to have more consistent range distributions, and our proposed method still demonstrates superior performance, proving its ability within the same depth range. 
% Note that most existing DFF method chose data with small range or re-scaling the existing depth to a smaller range mainly within $[0,3]m$, while our datasets are two times larger than theirs. 
Moreover, our network maintains fewer parameters than most competitors, thereby highlighting its exceptional effectiveness and efficiency.

In \cref{fig:result_vimeo}, we present our qualitative results compared to other state-of-the-art methods, demonstrating that the proposed EDFF consistently outperforms other models in detail and produces depth map closer to the ground truth.
% Note that the focal distances chosen by most frame-based methods are in a short range of $[0,3]m$, while ours vary from $[]$.

\subsection{Ablation Study}\label{Ablation Study}

To verify the effectiveness of the key components in our EDFF, we conducted an ablation study on the proposed FDCM and MDFB. For a fair comparison, we added convolutional layers to replace the removed modules, thus maintaining a consistent number of parameters. As shown in \cref{tab:ablation study}, we use the removal of both FDCM and MDFB as the baseline and add one module at a time. Without FDCM, the model experiences performance degradation, particularly on the EFS-NYUv2 dataset, which has an uneven depth range and is sensitive to focal distance information. This underscores the importance of FDCM in identifying the most relevant focal distance. Moreover, introducing MDFB improves performance by integrating the coarse depth from various levels of the decoder. With both FDCM and MDFB employed simultaneously, EDFF achieves the best performance.
\input{tables/ablation} 

\section{Conclusion}
In this letter, we propose a novel network, EDFF, which is designed to estimate sparse depth from event focal stacks. Utilizing the high temporal resolution introduced by the event focal stack, we implement attention and multi-scale coarse-to-fine mechanism to enhance depth estimation. Two synthetic EFS-based datasets are built for training and testing. Extensive experiments demonstrate that our method outperforms existing state-of-the-art approaches.

\noindent{\bf Limitation.} The proposed method cannot predict dense depth maps due to the inherent sparsity of events. Further research could explore combining traditional focal stacks with event focal stacks to compensate for the missing information.

\clearpage

\bibliographystyle{IEEEtran}
\IEEEtriggeratref{20}
\bibliography{IEEEfull,egbib}

\end{document}

%% file: tables/quantitaitve.tex
\begin{table*}[!t]
\centering
\renewcommand{\arraystretch}{1.1}
\caption{Quantitative comparisons on the EFS-NYUv2 and EFS-Blender datasets.}
\resizebox{\textwidth}{!}{
\begin{threeparttable}
\begin{tabular}{cccccccccccccc}
\hline
\multirow{2}{*}{Methods}&\multirow{2}{*}{Input} & \multicolumn{5}{c}{EFS-NYUv2} & & \multicolumn{5}{c}{EFS-Blender} & \multirow{2}{*}{\#Param.}  \\ 

% \cline{3-7} \cline{9-13}
& &RMSE$\downarrow$ & AbsRel$\downarrow$ & $\delta^1\uparrow$ & $\delta^2\uparrow$ & $\delta^3\uparrow$ & & RMSE$\downarrow$ & AbsRel$\downarrow$ & $\delta^1\uparrow$ & $\delta^2\uparrow$ & $\delta^3\uparrow$& 
\\ 
\hline
% VDFF& FS & 26.86 & 0.8237 & & & & & 19.48 & 0.5336 & & & & -- 
% \\
*DDFF \cite{hazirbas2019deep}& FS & 0.5211 & 0.1382  & 0.8266 & 0.9832 & 0.9982 & & 1.9006 & 0.3983 & 0.3721 & 0.6663& 0.8383 & 39.8M 
\\
*Defocus-Net \cite{maximov2020focus}& FS & \underline{0.1526} & \underline{0.0366}  & \underline{0.9888} & \underline{0.9989} & \underline{0.9998} & & 0.9105 & 0.1136 & 0.8783 & 0.9512 & 0.9761 & \textbf{3.7M} 
\\
AiFDepthNet \cite{wang2021bridging}& FS & -- & --  & -- & -- & -- & & 1.8013 & 0.3335 & 0.4688 & 0.7099 & 0.8795 & 16.5M \\
% \\
*DFF-DFV \cite{yang2022deep}& FS& 0.1677 & 0.0419 & 0.9807 & 0.9962 & 0.9996 & &\underline{0.8661} & \underline{0.1041}  & \underline{0.9012} & \underline{0.9571} & \underline{0.9778} &  19.5M \\
DDFS \cite{fujimura2023deep}& FS & 0.8347 & 0.2879 & 0.6343 & 0.8341 & 0.9201 & & -- & --  & -- & -- & -- &  26.3M \\
\hline
EDFF (Ours)& EFS & \textbf{0.0646}&\textbf{0.0137} & \textbf{0.9972} & \textbf{0.9996} & \textbf{0.9999} & & \textbf{0.8251}&\textbf{0.0869}  & \textbf{0.9087} & \textbf{0.9639} & \textbf{0.9832} &   \underline{6.2M}  
\\ 
\hline
\end{tabular}
\end{threeparttable}
}
\vspace{-1.7em}
\label{tab:quantative}
\end{table*}

%% file: tables/ablation.tex

\begin{table}[t]
\centering
\vspace{-0.3em}
\caption{The ablation experimental results of EDFF.}
\vspace{-0.3em}
\resizebox{0.98\linewidth}{!}{
\begin{tabular}{cccccccc}
\hline
\multirow{2}{*}{Models} & \multirow{2}{*}{FDCM}  & \multirow{2}{*}{MDFB} & \multicolumn{2}{c}{EFS-NYUv2}& \multicolumn{2}{c}{EFS-Blender} \\
                  &                &                  & \multicolumn{2}{c}{RMSE/AbsRel/$\delta^1$}          & \multicolumn{2}{c}{RMSE/AbsRel/$\delta^1$}        \\ \hline
Baseline            &          &                 &             \multicolumn{2}{c}{0.210/0.037/0.973}               &              \multicolumn{2}{c}{1.891/0.486/0.328}            \\
w/o FDCM                & \checkmark           &                   &               \multicolumn{2}{c}{0.177/0.029/0.983}               &              \multicolumn{2}{c}{0.847/\underline{0.091}/0.898}             \\
w/o MDFB               &            &      \checkmark             &               \multicolumn{2}{c}{\underline{0.075}/\underline{0.016}/\underline{0.996}}               &              \multicolumn{2}{c}{\underline{0.832}/\underline{0.091}/\underline{0.904}}             \\
w/ all              & \checkmark           &    \checkmark                 &     \multicolumn{2}{c}{\textbf{0.065}/\textbf{0.014}/\textbf{0.997}}          &      \multicolumn{2}{c}{\textbf{0.825}/\textbf{0.087}/\textbf{0.909}}        \\ \hline
\end{tabular}
}
\label{tab:ablation study}
\vspace{-1.5em}
\end{table}